\documentclass[letterpaper, 10 pt, conference]{ieeeconf}
\IEEEoverridecommandlockouts                            
\usepackage[T1]{fontenc} 
\usepackage{enumerate}
\usepackage{amsfonts}
\usepackage[colorinlistoftodos]{todonotes}
\usepackage[font=small,skip=12pt]{caption}
\usepackage{float}
\usepackage{subcaption}
\usepackage{bm}
\usepackage{booktabs}
\makeatletter
\let\NAT@parse\undefined
\makeatother
\usepackage{hyperref}
\usepackage{verbatim}
\usepackage{comment}
\usepackage{amsmath}
\usepackage{multirow}
\usepackage[noadjust]{cite}

\usepackage{mathtools}
\usepackage{dsfont}
\usepackage{tipa}
\usepackage{textcomp}
\usepackage[normalem]{ulem}
\DeclareMathAlphabet{\mathpzc}{OT1}{pzc}{m}{it}

\newcommand{\etal}{\textit{et al}.}
\hypersetup{
    colorlinks=true,
    linkcolor=red,
    filecolor=magenta,      
    urlcolor=magenta,
    pdftitle={Overleaf Example},
    pdfpagemode=FullScreen,
    }
\usepackage{graphicx}
\definecolor{ForestGreen}{rgb}{0.133, 0.545, 0.133}
\definecolor{CommentBlack}{rgb}{0.0, 0.0, 0.0}

\usepackage[ruled,vlined,linesnumbered]{algorithm2e}

\SetCommentSty{mycommfont}

\title{\LARGE \bf
Learning for \textcolor{CommentBlack}{Deformable Linear Object} Insertion Leveraging Flexibility Estimation from Visual Cues
}

\author{Mingen Li$^{1}$ and Changhyun Choi$^1$
\thanks{$^1$The authors are with the 
Department of Electrical and Computer Engineering, University of Minnesota, Minneapolis, MN 55455 {\tt\small \{li002852, cchoi\}@umn.edu}}%
\thanks{$^\dag$This work was supported in part by NSF Award 2143730 and MnDRIVE Initiative on Robotics, Sensors, and Advanced Manufacturing.}
}

\begin{document}
\maketitle 
\thispagestyle{empty}
\pagestyle{empty}

\begin{abstract}
Manipulation of \textcolor{CommentBlack}{deformable Linear objects} (DLOs), including iron wire, rubber, silk, and nylon rope, is ubiquitous in daily life.
These objects exhibit diverse physical properties, such as Young's modulus and bending stiffness. Such diversity poses challenges for developing generalized manipulation policies.
However, previous research limited their scope to single-material DLOs and engaged in time-consuming data collection for the state estimation. 
\textcolor{CommentBlack}{In this paper, we propose a two-stage manipulation approach consisting of a material property (e.g., flexibility) estimation and policy learning for DLO insertion with reinforcement learning.} Firstly, we design a flexibility estimation scheme that characterizes the properties of different types of \textcolor{CommentBlack}{DLOs}. The ground truth flexibility data is collected in simulation to train our flexibility estimation module. 
During the manipulation, the robot interacts with the DLOs to estimate flexibility by analyzing their visual configurations. 
Secondly, we train a policy conditioned on the estimated flexibility to perform challenging DLO insertion tasks.
Our pipeline trained with diverse insertion scenarios achieves an 85.6\% success rate in simulation and 66.67\% in real robot experiments. Please refer to our project page: 
\url{https://lmeee.github.io/DLOInsert/}

\end{abstract}

\section{Introduction}
\begin{figure}
\vspace{2mm}
   
    \begin{center}
    \includegraphics[width=0.47\textwidth]{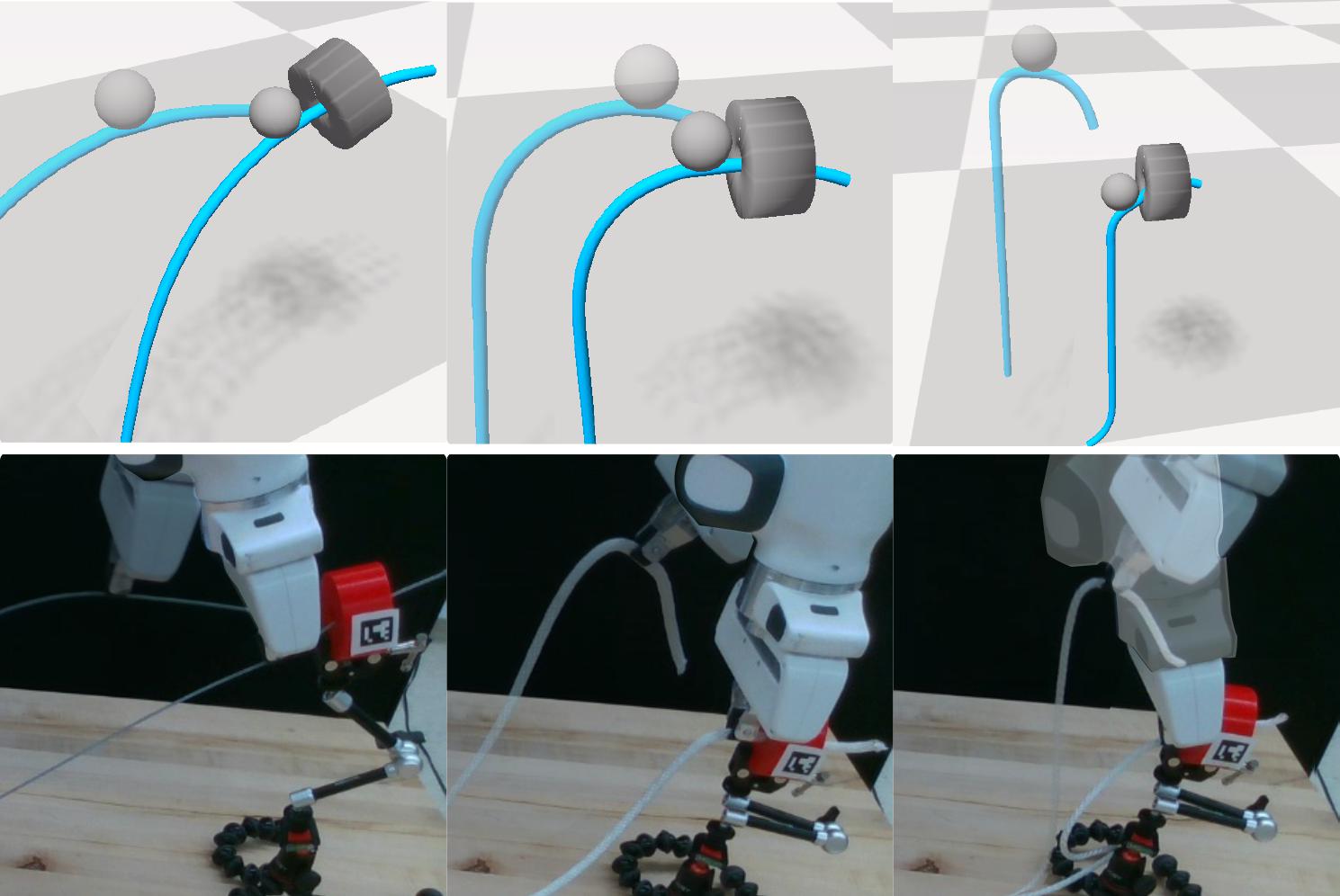} 
    \end{center}
    \vspace{-5mm}
    \caption{\textbf{Flexibility-aware DLO insertion.} The real DLO insertion demonstration (lower) and corresponding simulation motion (upper) with estimated flexibility of the real DLOs.}
    \label{fig:frontpage}
    \vspace{-4mm}
\end{figure}

\textcolor{CommentBlack}{Interacting with highly deformable objects is challenging due to the infinite number of degrees of freedom. Moreover, their physical properties vary significantly among different kinds of ropes.
Among manipulation tasks with deformable linear objects (DLOs), DLO insertion into a hole has extensive applications in healthcare, manufacturing, and households.} 
Imagine suturing a surgical incision with a deformable thread, routing cables inside an automobile frame, or inserting cables into a desk grommet.  
These tasks involve DLOs with vastly different physical properties, requiring different insertion strategies. 
\textcolor{CommentBlack}
{
For instance, the grasping point of a stiff iron wire differs significantly from that of a flexible nylon rope, where the latter requires a closer grip near the rope head for successful insertion (Fig.~\ref{fig:frontpage}). 
}
\textcolor{CommentBlack}{For successful insertion regardless of the DLO types, we have to address the following questions: (1) How to estimate the material properties (e.g., flexibility) from visual observations?,} (2) How to select the pick position for DLOs with different flexibility?, and (3) How should the agent generate a feasible manipulation trajectory to insert the DLO to a target?
Previous research \cite{lv2023samrl, LapGym2023} often fails to account for the diversity of DLO types. Lv~\etal \cite{lv2023samrl} rely on simplified rigid-body models that may yield inaccurate properties like elasticity and compliance. Additionally, these studies do not consider the various target hole configurations, such as different orientations and radii of holes, commonly encountered in real-world insertion tasks commonly encountered in real-world insertion tasks. Moreover, a heuristic design that the agent always grasps the tail-end, restricts the system from finding an optimal solution. 
\textcolor{CommentBlack}{To address these limitations, we propose a two-stage manipulation approach: the material property estimation of DLOs and an insertion policy learning conditioned on the estimated material property.} In the initial phase, we collect ground truth data regarding the DLO's state and flexibility via a single robotic interaction in a simulation and leverage a Graph Neural Network (GNN) to capture the underlying physics properties.
\textcolor{CommentBlack}{We leverage SoftGym \cite{corl2020softgym} as a simulator platform that uses position-based dynamics (PBD) as the backbone and offers fast and accurate modeling of dynamics of DLO and motion.}
Our flexibility estimation model is trained with ground truth data in the simulation. The estimated flexibility serves as a vital observation for reinforcement learning.
We employ a motion primitive for grasping the object and inserting it into a target, and the parameters of the motion primitive are optimized through a RL. In real-world experiments, a robotic system initiates by grasping a DLO and takes a single interaction step (grasping) to estimate its flexibility. It then executes the insertion task guided by the learned policy as shown in Fig.~\ref{fig:frontpage}.

The primary contributions of this paper are as follows:
\begin{itemize}
    \item We introduce a one-step flexibility estimation model for \textcolor{CommentBlack}{DLOs}.
    \item We propose a reinforcement learning framework for DLO grasping and insertion with the estimated flexibility.
    \item Our pipeline demonstrates its effectiveness in handling diverse DLOs, including iron wire, rubber, and nylon ropes in the real world without any further training for the RL policy.
\end{itemize}


\section{Related Work}
\subsection{DLO Manipulation}

DLO manipulation has been extensively studied with applications in robot-assisted surgery~\cite{tactileinsertion, Grannen2020untangling}, knot tying \cite{sundaresan2020learning,viswanath2023ltodo}, and cable handling \cite{she2020cable, yang2024attribute, wilson2023cable}. 
Some works employ supervised learning for state estimation using visual observation to address challenges like occlusion \cite{lv2023learning}, or dynamic DLO manipulation \cite{zhang2021robots}. 
Several works employ goal-conditioned techniques for rope configuration \cite{seita2023learning, yan2020selfsupervised}, or knot-tying \cite{sundaresan2020learning}. 
\cite{wang2019learning} utilizes a generative model to generate a sequence of transient images from goal observation. However, relying on predefined goals can restrict the system's adaptability and hinder its performance in dynamic scenarios.
Another approach \cite{li2023dexdeform} is to employ demonstration-based techniques to address DLO manipulation challenges, but this approach requires time-consuming data collection and struggles to generalize to diverse DLOs. Unlike prior works, our approach does not rely on demonstration data or prior knowledge of DLO properties, making it adaptable to a wide range of DLOs.

\textcolor{CommentBlack}{The DLO-in-Hole task has been studied for both needle threading  \cite{lv2023samrl, tactileGregorioInsertion, ropeinsertion2015} and assembly task \cite{galcblasm2022}.  \cite{ tactileinsertion} leverages tactile information for thread-tail finding and eyelet insertion. While these works rely on a limited range of material properties and typically assume tail-end grasping, our work can effectively estimate flexibility for various DLOs and approach optimal insertion.}

\subsection{Reinforcement Learning for Deformable Manipulation}
Reinforcement learning (RL) is frequently used in robotic manipulation for its adaptability and performance improvement in uncertain environments. 
As deformable simulators become more accessible \cite{corl2020softgym,gan2021threedworld, Li2020IPC}, researchers have introduced specialized frameworks for RL, such as AMBF-RL\cite{AMBF-RL}, SurRoL\cite{SurRoL}, dVRL\cite{dvrl2019}, and SoftGym\cite{corl2020softgym}.
Several studies have employed model-based RL to perform deformable manipulation tasks, including needle-threading \cite{lv2023samrl}, goal-conditioned deformable manipulation with video prediction \cite{ebert2018visual}, and 3D shape control of \textcolor{CommentBlack}{DLOs} \cite{olineyangshape}.
Lin~\etal~\cite{lin2019reinforcement} acquire DLO manipulation skills from RGB images using goal-reaching RL.
Singh~\etal~\cite{singh2023crisp} employ hierarchical RL for DLO manipulation through attainable sub-goals.
Yu~\etal~\cite{yu2022shape} use offline learning for the initial estimation of the DLO trajectory, followed by online updates to compensate for errors.
While these approaches obtain favorable results in deformable object manipulation, their generalization across diverse objects and scenarios requires further study. 
Our methods can directly transfer to real experiments with diverse environmental settings and a wide range of DLO types without further training or fine-tuning.




\begin{figure*}
   
    \begin{center}
    \includegraphics[width=0.8\textwidth]{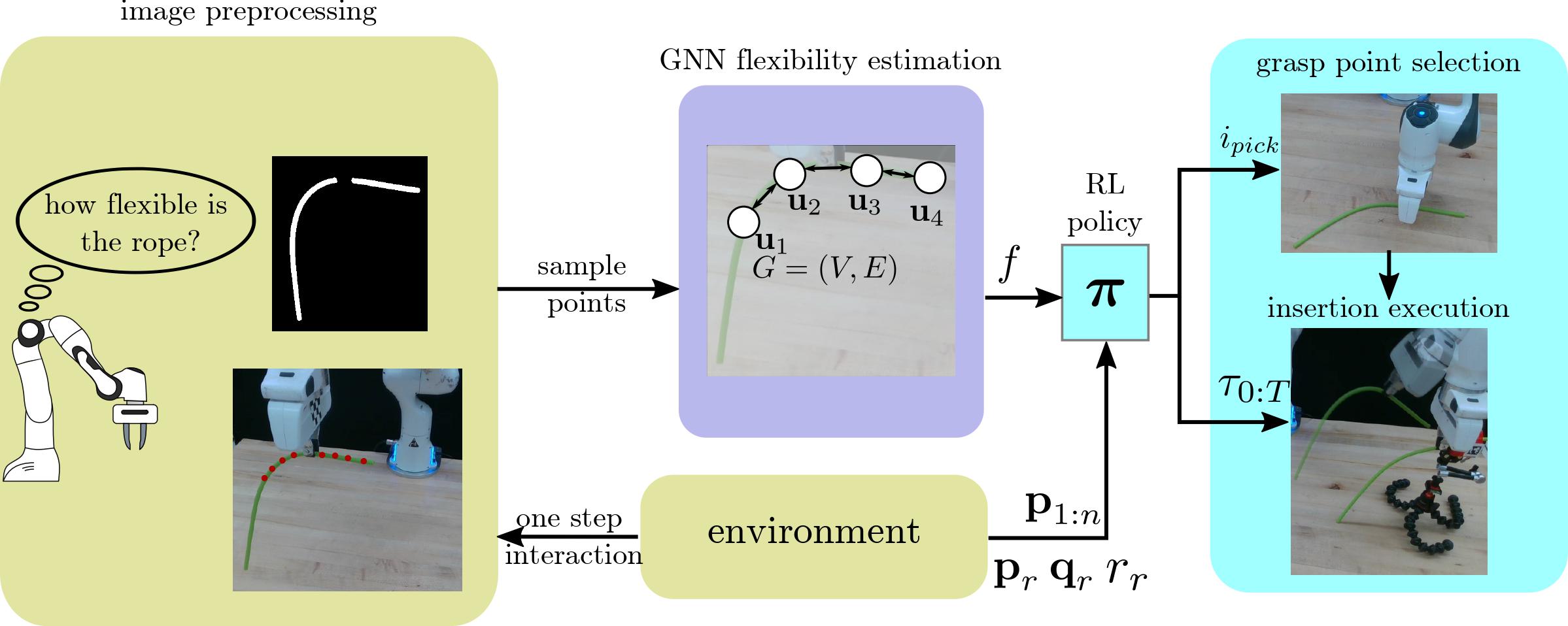} 
    \end{center}
    \vspace{-4mm}
    \caption{\textbf{\textcolor{CommentBlack}{Pipeline for flexibility-aware DLO insertion in real experiment.}} \textcolor{CommentBlack}{We start by asking the robot to grasp the testing DLO using a predefined pose and sample particles on the DLO with image prepossessing techniques. The DLO states are input to the GNN-based estimation module to predict flexibility.} Taking flexibility, DLO states and ring configuration as input, the policy can output a predicted grasping point and a control trajectory for insertion.}
    \label{fig:pipeline}
    \vspace{-3mm}
\end{figure*}

\section{Method}
\subsection{Problem Formulation}
In our task setup, the robot needs to insert a DLO into a target ring. 
We randomize ring properties (e.g. radius, position, and orientation) to improve the generalization of our model. In addition, we employ a deep ring hole, prompting the robot agent to consider a proper grasping point before insertion. \textcolor{CommentBlack}{To account for DLO variation, we introduce flexibility $f$, a material property closely associated with DLO manipulation. Flexibility is used to characterize a wide range of daily-life DLOs, as discussed in Section \ref{sec:defflex}}. \textcolor{CommentBlack}{Furthermore, we elaborate on the flexibility estimation in Section \ref{sec:flexest}.} 

We formulate the \textcolor{CommentBlack}{DLO} insertion task problem as a Markov Decision Process (MDP) $(\mathbf{S}, \mathbf{A}, r, \gamma)$. Other than properties of target rings, the state $\mathbf{S}$ consists of positions $\mathbf{p}_{1:n}$ of a DLO modeled with $n$ particles.
As for action space $\mathbf{A}$, we use a gripper with 3D translation motion and 1D rotation around one axis. We leverage a motion primitive for our grasp point selection and insertion, which generates long-horizon trajectories for the gripper as discussed in Section \ref{sec:rl_mopri}.


\begin{figure}
   
    \begin{center}
    \includegraphics[width=0.45\textwidth]{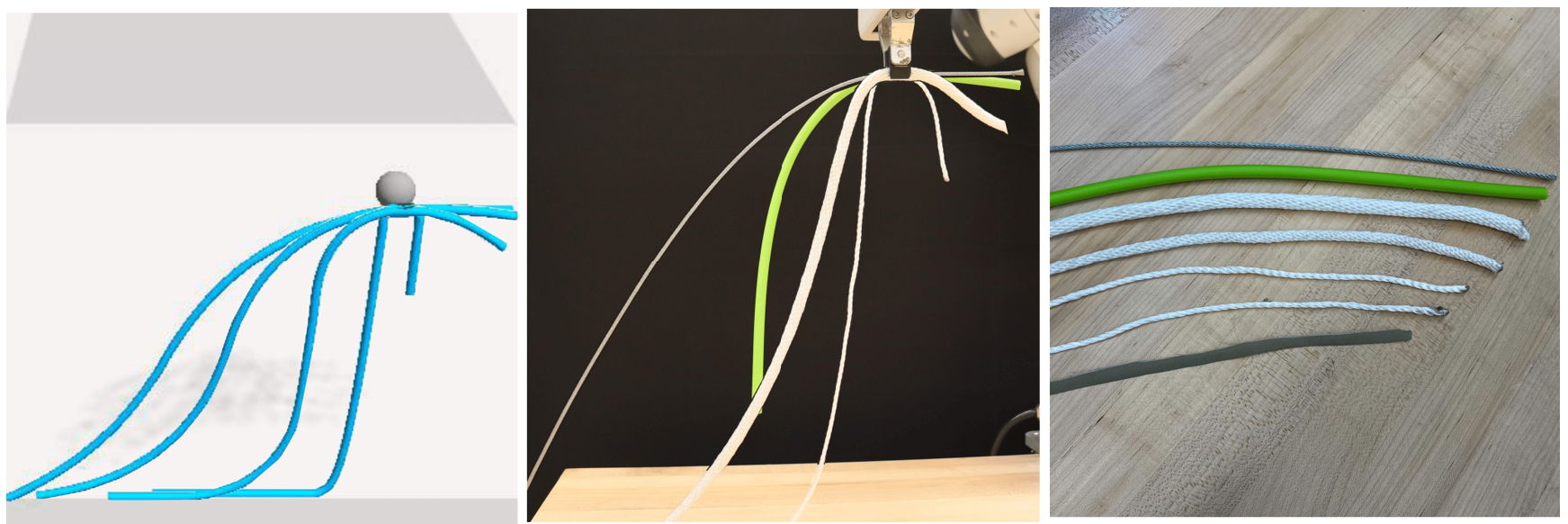} 
    \end{center}
    \vspace{-4mm}
    \caption{\textbf{Predefined grasping poses for DLO flexibility estimation} in simulation (left) and real experiment (middle). The right image shows the DLOs used for the real experiment.}
  
    \label{fig:multirope}
    \vspace{-3mm}

\end{figure}

\subsection{Defining Flexibility}
\label{sec:defflex}
To characterize various types of DLOs, we seek a property that encapsulates task-related information. Such a property should satisfy the following requirements: 
1) The property should be directly relevant to the insertion task, helping agents acquire valuable knowledge to improve training performance. 
2) It can be obtained or represented by visual observation.

 Before making contact with the ring, we noticed that the DLO could be treated as quasi-static under the influence of gravity.  This pre-insertion phase is crucial as insertion success depends on the relationship between the orientation of the target ring and the DLO's state $\mathbf{{p}}_{i_{p}:n}$ where $i_p$ is the index of the picked particle. This insight led to developing a property linked to the DLO's state under gravity, potentially enhancing policy training during insertion.
 
To simplify the property estimation, we assume the DLO always lies in the same plane perpendicular to the ground due to gravity. This plane has a normal vector $\hat{\mathbf{n}} = [x_n, y_n, 0]^T$, while the z-axis of the gripper coordinate system is $\hat{\mathbf{z}} = [0, 0, -1]^T$. We use Eq. (\ref{eq:transform}) to convert DLO particles $\mathbf{p}_i$ from world coordinates to the gripper coordinate system within the plane: 
\begin{equation}
\left(\begin{array}{c}
     x^\mathit{proj}_{i}  \\
     y^\mathit{proj}_{i}
\end{array} \right)= 
\left(\begin{array}{ccc}
     1&0&0  \\
     0&0&1
\end{array} \right)
\left( \begin{array} {c}
    (\hat{\mathbf{n}} \times \hat{\mathbf{z}})^T \\
   \midrule
    \hat{\mathbf{n}}^T \\
   \midrule
    \hat{\mathbf{z}}^T\\
\end{array}\right) \mathbf{p}_{i}.
\label{eq:transform}
\end{equation}
Building on these considerations and the concept of curvature \cite{curvature_book}, we define flexibility $f$ as shown below: 
\begin{equation}
\begin{split}
f = &-\frac {\frac{\Delta_y}{\Delta_x}(i_{p}+3, i_{p}+1) 
- \frac{\Delta_y}{\Delta_x}(i_{p}+1, i_{p})} {(x_{i_p+1}^\mathit{proj} - x_{i_p}^\mathit{proj})(1+ \frac{\Delta_y}{\Delta_x}(i_{p}+1, i_{p})^2)^{1.5}}
\end{split}
\label{eq:flexdef}
\end{equation}
where $\frac{\Delta_y}{\Delta_x}(a, b) = \frac{y_{a}^\mathit{proj} - y_{b}^\mathit{proj}}{x_{a}^\mathit{proj} - x_{b}^\mathit{proj}}$. 
Unlike curvature, flexibility can take negative values, indicating a highly stiff DLO.

As shown in Fig. \ref{fig:multirope}, we have observed that, when consistently applying the same grasping pose to the DLO, it consistently and uniformly maintains a static bending configuration. This phenomenon remains consistent whether the DLO is rigid or exceedingly flexible and deformable. To capture such subtle deformations at specific points along the DLO, we employ the concept of curvature. Although curvature represents sharpness at a singular focal point, it proves adequate in our ability to devise a consistent manipulation approach for characterizing the bending configuration of the entire length of the DLO. The discrete version of the curvature calculation is shown in Eq. (\ref{eq:flexdef}).

\subsection{GNN-based Flexibility Estimation}
\label{sec:flexest}
After defining flexibility, we use a Graph Neural Network (GNN)\cite{kipf2017gcn} to estimate the DLO flexibility. A graph $G= (V, E)$ is constructed naturally by using sampled particles as vertices $V$ and directly connecting adjacent particles as edges $E$, forming a representation for DLO. Specifically, each particle becomes a vertex: $v_i = [\mathbf{p}_{i}]$, and adjacent vertices are connected with undirected edges: $E=\{(v_i, v_j)|~|i - j| = 1\}$. We employ a four-layer message-passing network and a linear layer to predict flexibility values. We augment the collected simulation flexibility data to enhance data diversity and improve performance in noisy real-world scenarios. This involves introducing a scaling factor $\lambda \in [0.85, 1.0]$ and adding translational Gaussian noise $\mathbf{n}_{1:n} \sim \mathcal{N}(0, 0.003)$ to $\mathbf{p}_{1:n}$, resulting in modified positions $\mathbf{p}'_{1:n} = \lambda (\mathbf{p}_{1:n}+ \mathbf{n}_{1:n})$\cite{shao2022transformer}.

\subsection{Reinforcement Learning with Motion Primitive} \label{sec:rl_mopri}
\begin{figure}[h]
    \begin{center}
    \includegraphics[width=0.3\textwidth]{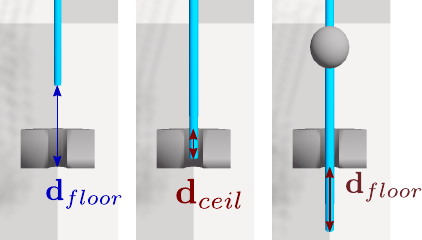} 
    \end{center}
    \vspace{-5mm}
    \caption{\textbf{Distance reward visualization.} $\mathbf{d}_{floor}$ contributes negatively to reward when the DLO is not inserted. After insertion, a positive distance reward is given when the DLO is halfway or completely through.}
    \label{fig:reward}
\end{figure}
This subsection describes details of the motion primitive and an RL method for insertion conditioned on the previously acquired flexibility estimation.
\textbf{Task Definition}: Insertion involves the robot inserting different types of DLO through a ring until the DLO tip emerges on the other side, as depicted in Fig.~\ref{fig:reward}.
\textbf{Observation Space}:
The observation space contains the position of DLO particles, $\mathbf{p}_{1:n}$, ring's position $\mathbf{p}_{r}$, angle $\mathbf{q}_r$, and radius $ r_r$. The ground truth flexibility value of the DLO $f$ is optionally provided depending on the baseline setting.
\textbf{Motion Primitives and Action Space}:
In the SoftGym environment, we use a gripper with position $\mathbf{p}_p^t$ and orientation $\mathbf{q}_p^t$ to enable the interaction with the DLO. To perform long-horizon planning, we leverage a motion primitive that takes grasped particle index $i_{p}$, starting pose $(\mathbf{p}_p^{0}, \mathbf{q}_p^{0})$ and ending pose $(\mathbf{p}_p^{T}, \mathbf{q}_p^{T})$ as input in Fig. \ref{fig:actspace}. The output trajectory includes DLO grasping and insertion motion. A customized PD controller is utilized to generate the control trajectory $\boldsymbol{\tau}_{1:T}$ for the insertion.
The insertion is conducted in a planar plane formed by the DLO and the target ring. The agent observes the environment only at the beginning of the episode, and the insertion is executed in an open-loop manner.
\textbf{Policy Training}: We use Soft Actor-Critic (SAC) \cite{sac2018} for training with MLP for both actor and critic networks. We train with three random seeds using the default parameters from in \href{https://github.com/DLR-RM/stable-baselines3}{stable-baselines3}.
\textbf{Reward Function}:
As shown in Fig.~\ref{fig:reward}, we define the reward function for three distinct stages: the initial stage (left), the halfway point (middle), and the successful passage (right) of the DLO through the ring. These stages are delineated by binary values of the $\mathit{rope\_in}$ and the $\mathit{rope\_out}$ variables. Eq. (\ref{eq:reward}) incorporates both a stage-based reward component and a penalty term $r_\mathit{pen}$ to account for abnormal stretching of the elastic DLOs during simulation:
\begin{equation}
\begin{split}
&r=0.5 (\mathit{rope\_in}+\mathit{rope\_out})+r_\mathit{pen}+ r_\mathit{dist}\\
& r_\mathit{dist} = 
\begin{cases}
-10 \cdot d_\mathit{floor} & \text{if not $\mathit{rope\_in}$}\\
5 \cdot d_\mathit{ceil} & \text{if $\mathit{rope\_in}$ and not $\mathit{rope\_out}$}\\
10 \cdot d_\mathit{floor} & \text{if $\mathit{rope\_out}$}\\
\end{cases}
\end{split}
\label{eq:reward}
\end{equation}
If the DLO segments are stretched to 120\% of the original length during simulation, it is considered abnormal and the penalty $r_\mathit{pen}=-2$ is applied.

    





\begin{table*}
\begin{center}
 \scalebox{1.0}{
 \centering
 \begin{tabular}{ccccccc}
 \toprule
    Method &  Evaluation  &  Ours w/ $f$ & Ours w/o $f$ & VB w/ $f$ & VB w/o $f$  & Random $f$  \\
      \midrule
   \multirow{2}{*}{rand $\theta$} &Success Rate (\%) $\uparrow$ & $\bm{78}\pm\bm{2}$& $56\pm1$ & $40$ & $32$ & $21\pm 4$  \\
   & Avg. dis. (cm) $\uparrow$ & $\bm{3.92}\pm\bm{0.51}$ & $0.00\pm0.16$ & $-0.98$ & $-0.12$ & $-5.35\pm0.48$  \\
   \midrule

   \multirow{2}{*}{fix $\theta$}& Success Rate  (\%) $\uparrow$  & $\bm{71}\pm\bm{1}$ & $39\pm5$& $31$ &$38$ & $10\pm3.0$ \\
   &Avg. dis. (cm) $\uparrow$ & $\bm{2.00}\pm\bm{0.46}$ & $-2.72\pm0.79$& $-1.56$ &$-1.42$ & $-7.00\pm0.44$\\
   \bottomrule
   
 \end{tabular}
 }
 \end{center}
  \vspace{-4mm}
 
 \caption{\textbf{Evaluation of all baselines in simulation.} For baselines trained with ring angle variation, random angle values are assigned for environments with different flexibility to perform evaluation. The ring angle is always $\pi/2$ rad for the fixed $\theta$. The ring radius used for evaluation is 1cm.}
\label{tab:evalall}
\vspace{-3mm}

\end{table*}

\section{Experiment and Result}

In this section, we detail an experiment about flexibility estimation and how flexibility can affect RL policy training and insertion task competence.

\subsection{Experiment Detail}
\subsubsection{Flexibility Estimation} In this part, we illustrate our flexibility estimation functions and baselines.

\noindent
\textbf{Data Collection}: To simulate DLOs with extreme flexibility value, we adopt the DLO configuration in SoftGym \cite{corl2020softgym} built upon the PyFleX \cite{pyflex2014}. Specifically, we modify DLO stiffness by adding extra springs, reducing the constraint solver iterations, and decreasing bending stiffness. To obtain ground truth flexibility data, we grasped the middle of DLOs (particle number $n=20$) and recorded their stationary state positions using Eq. (\ref{eq:flexdef}). We collected a consistent dataset with linearly sampled DLO flexibilities.

\noindent
\textbf{Data Augmentation}: With our labeled flexibility DLO data, we aim to capture more realistic DLO interactions to estimate flexibility. To address variations in DLO length, we introduce an unbalanced state by using a longer DLO ($n=40$) and biasing the grasping point to $i_{p}=10$, as depicted in Fig. \ref{fig:multirope} (left). We generate a dataset for training the flexibility estimation module by augmenting part of the state $\mathbf{p}_{21:40}$ with noise and pairing it with flexibility values.

\noindent
\textbf{Evaluation}: To evaluate flexibility estimation with actual DLOs, we replicate the same DLO interaction demonstrated in Fig. \ref{fig:multirope} (middle). We create a DLO mask using SAM~\cite{kirillov2023segany}, unproject the center-line onto a predefined gripper coordinate plane to sample DLO particles $\hat{\mathbf{p}}_{21:40}$. We sample particles with a distance of 1.2cm,  corresponding to the rest length of the DLO segment in SoftGym. Our evaluation dataset comprises 31 frames from 7 different DLOs. The simulated DLO with the closest estimated flexibility $\mathbf{p}_{21:40}$ is used for evaluation. Performance is assessed using the point-point distance described below:
\begin{equation}
d_{pp} = \frac{1}{20}\sum_{i=21}^{40} | \mathbf{p}_i - \hat{\mathbf{p}}_i |.
\label{eq: pp distance}
\end{equation}

\noindent
\textbf{Baselines}: We evaluate flexibility estimation using various baselines:
\begin{itemize}
    \item \textcolor{CommentBlack}{Analytic: The analytical method computes the theoretical flexibility using Eq. (\ref{eq:flexdef}) and the state extracted from real images.}
    \item Multi-Layer Perceptron (MLP): MLP has two linear layers with a hidden dimension of 8.
    \item GNN without data augmentation: Since GNN yields the best performance among all baselines, we test GNN without data augmentation to demonstrate its effectiveness of it.
\end{itemize}





   
   
  



\begin{figure}[t]
    \begin{center}
    \includegraphics[width=0.6\linewidth]{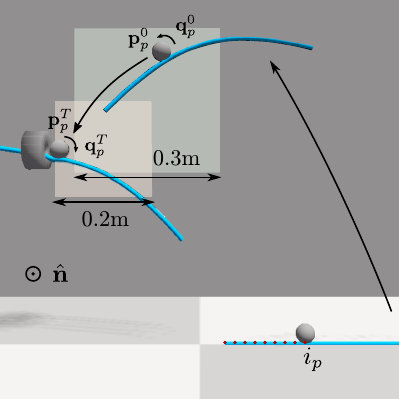} 
    \end{center}
    \vspace{-5mm}
    \caption{\textbf{Action space visualization.} The upper right square shows the range of $\mathbf{p}_{p}^{0}$ and the lower left square shows the range of $\mathbf{p}_{p}^{T}$. The rotation axis $\hat{\mathbf{n}}$ is pointing outward.}
    \label{fig:actspace}
\vspace{-5mm}
\end{figure}
\subsubsection{Simulation Setup}

In both simulation and real experiments, we assume insertion takes place within the two-dimensional plane, resulting in a 7-dimensional action space ($i_{p}$, $\mathbf{p}_{p}^{0},\mathbf{p}_{p}^{T} \in \mathbb{R}^2$, $\mathbf{q}_p^0$, $\mathbf{q}_p^T$) as shown in Fig. \ref{fig:actspace}. In practice, we implement $\mathbf{q}_p^0$, $\mathbf{q}_p^T$ as rotation with respect to axis $\hat{\mathbf{n}}$. Since the agent knows the ring's position and orientation, the action space is defined as a rectangle relative to the ring's local coordinates to ensure fast and informed exploration during RL.

\noindent
\textbf{Randomization}: We randomize the position of the ring within a 10cm square and the initial DLO position within 10cm inside the plane. We also vary the ring radii from challenging small (1cm) to easier large (2.5cm) sizes, and the ring angle $\theta$ within the range of $[0, \frac{3}{4}\pi]~\text{rad}$, covering positions from facing upward to a $45\deg$ downward angle.  


\noindent
\textbf{Baselines}: We conduct various baseline experiments for our simulation:
\begin{itemize}
    \item Fixed angle (fix $\theta$) and ring angle variation (rand $\theta$): Fixed angle variation provides the RL policy with one fixed challenging angle environment ($\theta=\pi/2 \ \text{rad}$), while ring angle variation ($\theta$) provide the target ring angle change within a limit ($\theta \in [0, 3\pi/4] \ \text{rad}$).
    \item Flexibility observation (Ours w/wo $f$): Whether providing estimated flexibility as an observation or not.
    \item Random flexibility (Random $f$): Training the agent with flexibility but introducing a random flexibility value during evaluation.
    \item Visual baseline without flexibility (VB w/o $f$): A heuristic method where the DLO is grasped at point $i_p=5$ and held horizontally. 
    We calculate the angle formed by the ring, gripping point, and DLO tip and subsequently perform a rotation using this angle, assuming that it will align the DLO in parallel with the ring hole.
    \item  Visual baseline with flexibility (VB w/ $f)$: Similar to the previous baseline, but introducing flexibility to grasp particle $i_p$ using a square root function.
\end{itemize}

\begin{figure}[th]
    \begin{center}
         \includegraphics[width=0.45\textwidth]{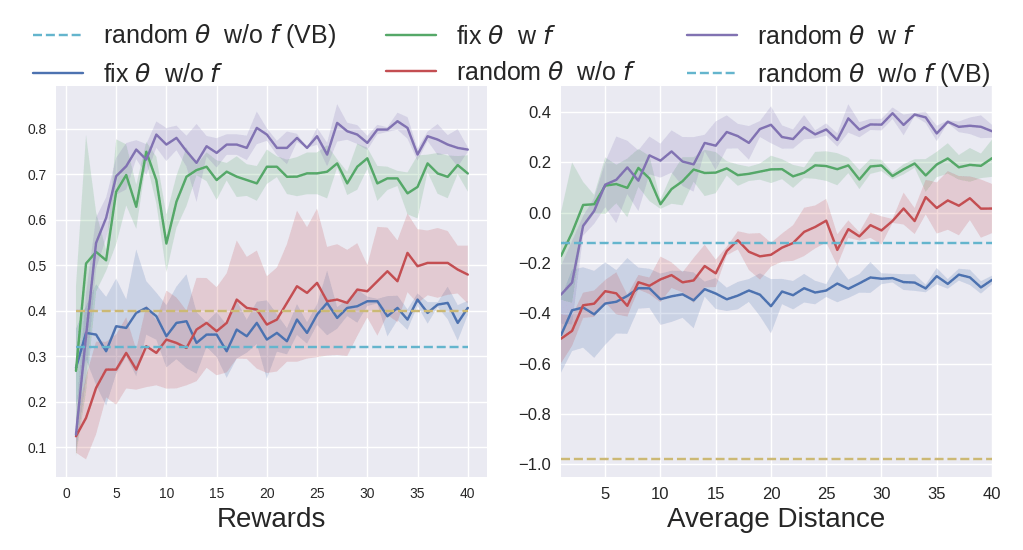}
    \end{center}
         
         \label{fig:training_curve}
    
\vspace{-5mm}

     \caption{\textbf{The training curves (left) and the average distance (right).} Average distance (avg. dis.) is the average signed endpoint distance. Similar to the reward definition, it equals to $d_\mathit{floor}$ when $\mathit{rope\_out}$ and $-d_\mathit{floor}$ otherwise.}
     \label{fig:traincuv}
\end{figure}

\noindent
\textbf{Training Details}: We have 4 sets of training settings (fix or random $\theta$, Ours w/wo $f$). 
We train each of them with 3 random seeds, totaling 40,000 episodes. We only have a 40k state-action pair for dynamics primitive and the policy training is converged (Fig. \ref{fig:traincuv}). In our evaluation, successful insertion is defined as the DLO penetrating the ring's floor (Fig.~\ref{fig:reward}). We evaluate the performance by calculating the endpoint distance based on $d_\mathit{floor}$. Prior to success, the distance is assigned as negative ($-d_\mathit{floor}$), while a high endpoint distance indicates strong insertion performance.

\subsection{Simulation Result Analysis}

 In this part, we discuss how our flexibility estimation and motion primitive affects training performance. 

\begin{table}[th]
\begin{center}
\scalebox{1.0}{
\begin{tabular}{cc} 
\toprule
model & $d_{pp}$ (mm)  \\ 
\midrule
Analytic & $11.80 \pm 13.52$\\
MLP  & $9.69 \pm 12.01$\\
GNN w/o aug &  $9.76 \pm 12.7$\\
GNN (Ours) & $\bm{9.16}\pm\bm{11.6}$\\

\bottomrule
\end{tabular}
}
\end{center}
\vspace{-4mm}

\caption{\textcolor{CommentBlack}{\textbf{Evaluation result for using different backbone or the analytical method for flexibility estimation model.} GNN achieves the smallest point-point distance and the smallest standard deviation.}}
\label{tab: flexeval}
\end{table}

\begin{figure}[thb]
    \begin{center}
    \includegraphics[width=0.45\textwidth]{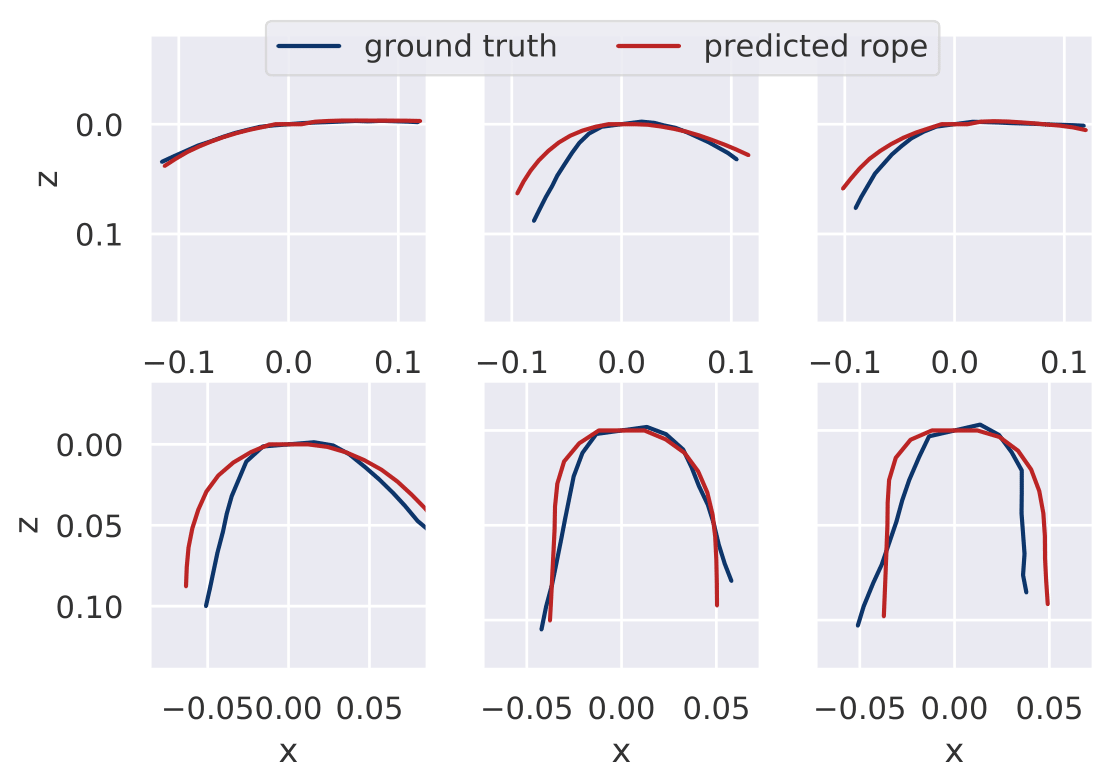} 
    \end{center}
\vspace{-6mm}
   
    \caption{\textbf{Visualization of real DLO flexibility estimation.} We estimate flexibility for the real DLOs and compare  with the simulated DLOs which have the closest flexibility.}
  
    \label{fig:flexestrope}
\vspace{-5mm}
\end{figure}

\begin{table*}[t]
\begin{center}
    
\scalebox{1.0}{
\begin{tabular}{cccccccc} 
\toprule
\multirow{2}{*}{Methods} & \multirow{2}{*}{Ring angle (deg)} & \multicolumn{2}{c}{$\approx$66}&\multicolumn{2}{c}{$\approx$90} &\multicolumn{2}{c}{$\approx$110}  \\ 
\cmidrule(lr){3-4}\cmidrule(lr){5-6}\cmidrule(lr){7-8}
&  & Success (\%)$\uparrow$ & Avg. Dis. (cm)\textcolor{CommentBlack}{$\uparrow$} & Success (\%)$\uparrow$ & Avg. Dis. (cm)\textcolor{CommentBlack}{$\uparrow$} & Success (\%)$\uparrow$ & Avg. Dis. (cm)\textcolor{CommentBlack}{$\uparrow$} \\
\midrule
VB w/ $f$&rand $\theta$& 29& 8.10& 29& 3.60&  0& 0.00\\
\multirow{2}{*}{Ours w/ $f$}
& fix  $\theta$ & -- & -- & 57 & 2.98 & -- & -- \\
&rand $\theta$ &  86 & 2.21 & 86 & 2.04 &29&  1.71\\
\midrule

\end{tabular}}
\end{center}
\vspace{-6mm}
\caption{\textbf{Real experiment results for insertion.} The ring angle above is an approximated value from image perception. Perception noise and potential robot collision may cause ring angle deviation during the experiment.}
\label{tab:insertrealeval}
\vspace{-5mm}
\end{table*}

\noindent
\textbf{Accuracy of Flexibility Estimation}: The baselines' evaluation results are summarized in Table \ref{tab: flexeval}. 
\textcolor{CommentBlack}{Directly calculating flexibility in Eq. (\ref{eq:flexdef}) with local DLO state only is not accurate due to noise in real experiments. Also, it fails to consider the complete DLO states which offer more informative insight into flexibility.} Table \ref{tab: flexeval} highlights GNN's superior performance with the smallest average distance and standard deviation, showing the strength of the graph-based neural network. In Fig. \ref{fig:flexestrope}, we compare the shapes of the estimated and real DLOs in the gripper's plane. 

\noindent
\textbf{Necessity of Flexibility Estimation}: To explore the necessity of flexibility estimation, we compared RL training with and without flexibility in Table \ref{tab:evalall}. Compared with training without flexibility, Our w/ $f$ obtained a 22\% and 32\% higher success rate on average for fix $\theta$ and rand $\theta$, respectively. The substantial gap underscores that the policy without flexibility is underperforming to the diverse DLOs in our environment. Furthermore, our approach achieves a significantly larger average distance, suggesting that it is closer to obtaining the optimal solution compared to cases without flexibility. Random $f$ shows the worst performance among all the methods because of significant divergence in optimal grasping points and insertion motion primitives for DLOs with various flexibility. These divergences prevent the achievement of uniformly superior outcomes in various flexibility situations.
Consequently, without proper consideration of flexibility, achieving the optimal insertion actions for different DLO types is unachievable.

\begin{figure}[ht]
    \begin{center}
    \includegraphics[width=0.8\linewidth]{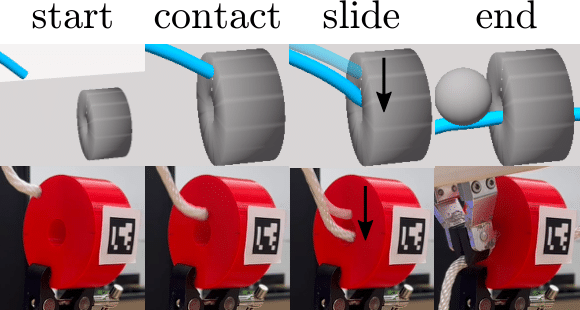} 
    \end{center}
    \vspace{-5mm}
    \caption{\textbf{Inserting by sliding in simulation (upper) and real experiment (lower).} The DLO tip first contacts with the upper ring wall and slides into the ring.}
    \vspace{-3mm}
    \label{fig:slideinsert}
\end{figure}

\noindent
\textbf{Comparing Against the Visual Baselines}: As shown in Table \ref{tab:evalall}, our method with flexibility (Ours w/ $f$) outperforms the visual baselines (VB w/ $f$) with a large margin. This is because our mapping between flexibility and grasp point selection is not as trivial as a simple function, but may involve higher-order terms. Moreover, the conventional approach of parallel insertion is sub-optimal due to the deformable nature of the objects and fails once DLOs collide with the ring wall. \textcolor{CommentBlack}{DLOs}, unlike rigid objects with well-defined transformations, tend to deform during motion, contradicting the initial assumption of parallel alignment. Additionally, parallel insertion often leads to head-on collisions with the thick ring wall, rendering contact-based insertion impossible. In contrast, our method tilts the DLO slightly, as illustrated in Fig. \ref{fig:slideinsert}, increasing the likelihood of a successful insertion.

\begin{figure}
    \begin{center}
    \includegraphics[width=0.5\textwidth]{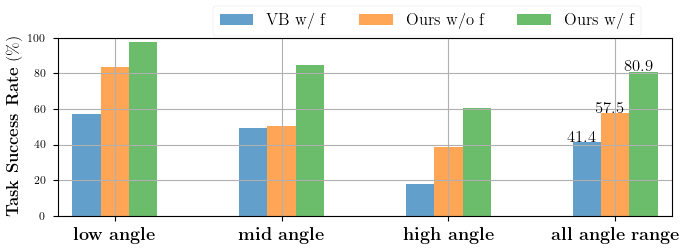} 
    \end{center}
    \vspace{-5mm}
    \caption{\textbf{Evaluation of baselines with different ranges of target ring angles in simulation.} Low-angle $\theta\in[0,\pi/4] \ \text{rad}$, mid-angle $\theta\in[\pi/4,\pi/2] \ \text{rad}$, high-angle $\theta\in[\pi/2,3\pi/4] \ \text{rad}$.} 
    \label{fig: randevalaangledifferent}
\vspace{-4mm}
\end{figure}

\noindent
\textbf{Comparing Performance Under Different Target Ring Angles}
We evaluate how the target ring angle influences the insertion task using a different ring angle range as shown in Fig \ref{fig: randevalaangledifferent}, the task success rate drops as the ring angle increases, indicating that the task difficulty is dependent on the ring angle. Nevertheless, our method achieves the highest level of success among all tested scenarios, demonstrating its proficiency in addressing the manipulation challenges posed by varying ring angles.



\subsection{Real Robot Evaluation}
We also conducted real experiments to verify the capability of our model in the real world.

\noindent\textbf{Experiment Setup}:
 We employed an Intel Real Sense camera to capture RGB images of the DLO and determined the orientation and position of the ring using Aruco markers\cite{garrido2014aruco}. Our manipulation was carried out using a Franka Emika robot with custom fingertips for DLO handling. For DLOs with limited depth information due to noise in the depth channel, we conducted flexibility estimation interactions at the plane positioned at $x=0.35m$ and unprojected this data using the camera transformation matrix. The insertion motion occurred at the plane corresponding to the ring's position. 
 \textcolor{CommentBlack}{We examined our method with 7 different testing DLOs in real experiment evaluation, including wire rope, rubber rope, thick nylon, median nylon, thin nylon, soaked thin nylon, and silk rope. We conducted 14 experiments for fix $\theta$ and 21 experiments for rand $\theta$.} The target ring had the same configuration as the one in the simulation, having an inner radius of 1cm, ring depth of 4cm, and outer radius of 4. We tested with three different ring angles.

\noindent
\textbf{Result Analysis}: Table \ref{tab:insertrealeval} summarizes the real experiment performance, and visualization of real DLO insertion is shown in Fig.~\ref{fig:frontpage}. We observe a sharp success rate drop as the target ring angle increases from 90 deg to 110 deg, which is consistent with our findings in the simulation. Our model trained with random $\theta$ achieves the best result and outperforms that trained with fix $\theta$ where the ring angle is 90 deg. 
While our method trained in a ring angle-variant environment adapts to different angles (66 and 110 deg), our method trained with fix $\theta$ overfits to a single ring environment and hence is more sensitive to ring angle deviation.
The VB has the worst success rate, characterized by frequent head-on collisions. The success heavily depends on the rigid characteristics of the object. The wire rope plays a pivotal role for VB and consistently achieves the longest average distance compared to other methods. Furthermore, VB is also significantly influenced by ring angle detection.

\section{Conclusion}
We presented a general robotic insertion task encompassing diverse DLOs and ring configurations. 
To accomplish this, we defined flexibility as a property that can be visually assessed and that guides the insertion task. We found that our flexibility-aware policy outperforms those without flexibility and could find a reliable trajectories than the considered baselines. Additionally, our policy exhibited good sim2real performance for various real DLO types. 
\clearpage
\newpage
\bibliographystyle{ieeetran}
\bibliography{varrope}

\end{document}